\documentclass[10pt,twocolumn,letterpaper]{article}

\usepackage{cvpr}
\usepackage{times}
\usepackage{epsfig}
\usepackage{graphicx}
\usepackage{amsmath}
\usepackage{amssymb}
\usepackage{subfigure}

\usepackage[pagebackref=true,breaklinks=true,letterpaper=true,colorlinks,bookmarks=false]{hyperref}

\cvprfinalcopy % *** Uncomment this line for the final submission

\def\cvprPaperID{2141} % *** Enter the CVPR Paper ID here

\ifcvprfinal\fi
\begin{document}

%%%%%%%%% TITLE
\title{Self-adaptive Single and Multi-illuminant Estimation Framework \\
based on Deep Learning}

\author{Yongjie Liu, Sijie Shen
% Jiemin Zhou, Takayoshi Kawaguchi \\
% DJI Innovations Inc. \\
% Tokyo, Japan \\
% {\tt\small \{yongjie.liu, sijie.shen, jiemin.zhou, takayoshi.kawaguchi\}@dji.com}
% \and
}

\maketitle
%\thispagestyle{empty}

%%%%%%%%% ABSTRACT
\begin{abstract}
  Illuminant estimation plays a key role in digital camera pipeline system,
  it aims at reducing color casting effect due to the influence of non-white illuminant.
  Recent researches~\cite{shi2016deep},~\cite{hu2017fc} handle this task by
  using Convolution Neural Network (CNN) as a mapping function from input image to a single illumination vector.
  However, global mapping approaches are difficult to deal with scenes under multi-light-sources.
  In this paper, we proposed a self-adaptive single and multi-illuminant estimation framework,
  which includes the following novelties:
  (1) Learning local self-adaptive kernels from the entire image for illuminant estimation with encoder-decoder CNN structure;
  (2) Providing confidence measurement for the prediction;
  (3) Clustering-based iterative fitting for computing single and multi-illumination vectors.
  The proposed global-to-local aggregation is able to predict multi-illuminant regionally by utilizing
  global information instead of training in patches~\cite{bianco2015single}~\cite{gijsenij2012color},
  as well as brings significant improvement for single illuminant estimation.
  We outperform the state-of-the-art methods on standard benchmarks with the largest relative improvement of 16\%.
  In addition, we collect a dataset contains over 13k images for illuminant estimation and evaluation.
  The code and dataset is available on \url{https://github.com/LiamLYJ/KPF_WB}. 
\end{abstract}

\begin{figure}[t]
\begin{center}
% \fbox{\rule{0pt}{3in} \rule{0.9\linewidth}{0pt}}
\footnotesize
\begin{tabular}{cccccc}
\hspace{0.3cm} Input & \hspace{0.15cm} Clustering & Local Fitting & \hspace{0.25cm} GT & \hspace{0.45cm} FC4~\cite{hu2017fc} \\
\end{tabular}
\includegraphics[width=\linewidth]{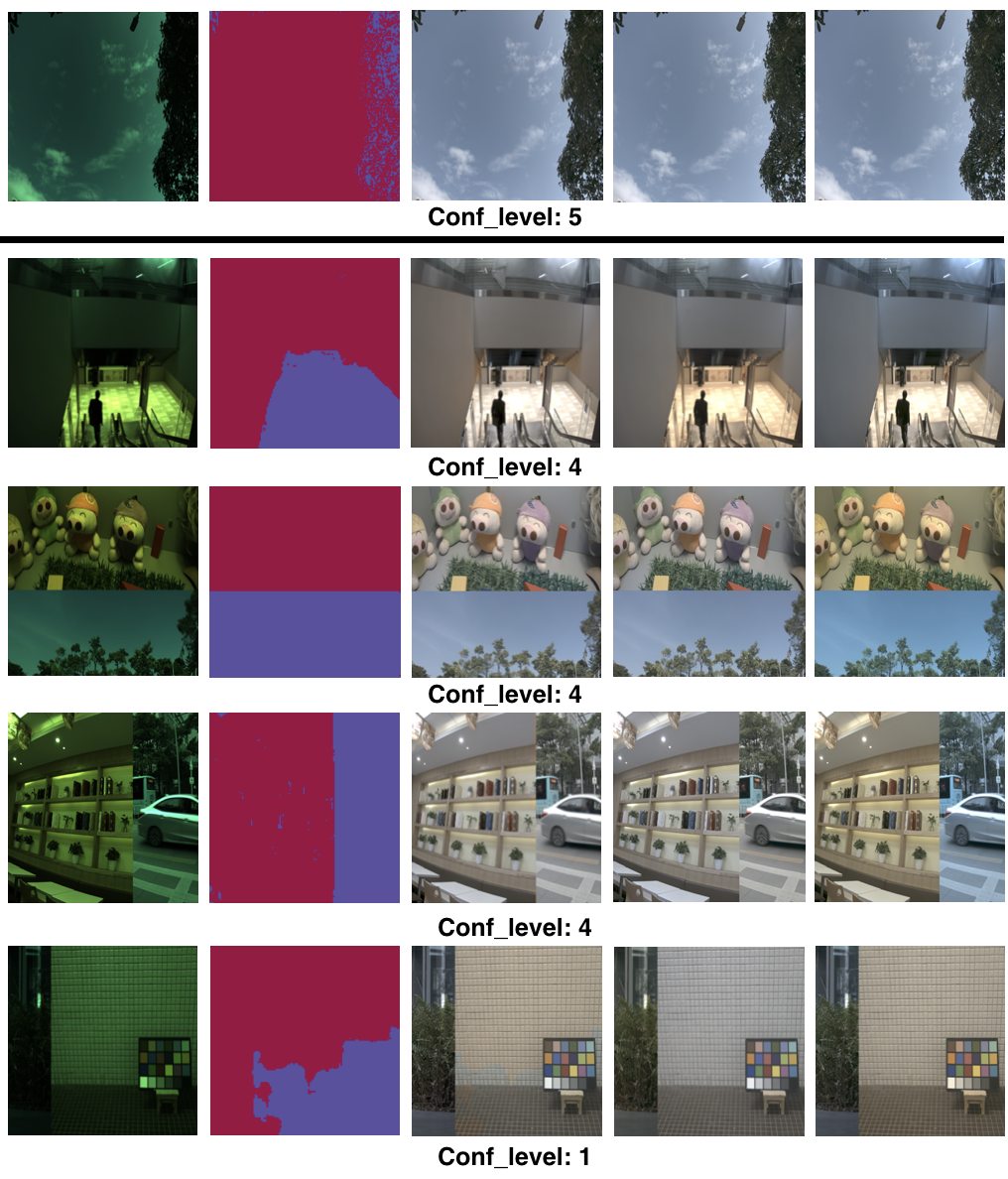}
\end{center}
   \caption{Illuminant estimation results of our proposed method.
            \textbf{top:} Results under single light source.
            \textbf{down:} Results under two distinct light sources (real and synthesized).
            The results of our proposed method are shown in the collum of clustering and local fitting,
            confidence levels are given below, the larger, the better.
            See Section~\ref{section_framework} for more details.
            GT denotes for ``ground truth".
            The results of the state-of-art CNN-based method FC4 is based on
            \setcounter{footnote}{0}
            the author's released code~\protect\footnotemark.
            Images are gamma-corrected ($\gamma = 1/2.2$) for display.}
\label{fig:show_off}
\end{figure}
\footnotetext{https://github.com/yuanming-hu/fc4}

\section{Introduction}
% What is color constancy, and why we need illuminant estimation in detail.

In recent years, there has been a considerable growth in digital cameras market.
Illuminant estimation is one of the core factors to improve the quality of captured photographs.
It is responsible for adjustment of the intensities of colors,
typically in primary colors: red, green and blue.
The goal of illuminant estimation is to render the neutral colors (the colors under white light source) correctly under non-white light sources.
Hence, general method in a camera pipeline system has an alias of ``White Balance" (WB).
Apart from photography, a lot of computer vision tasks are based on extracting comprehensive information from
the intrinsic colors of objects, \textit{e.g.} image segmentation, object recognition, \textit{etc}.

% method on real image pipeline and limitations
Unlike Human Visual System (HVS), which can automatically compensate for the illumination effect
of color perception by adjusting the photoreceptors with our brain,
a digital camera system needs to estimate the environment illuminant
accurately in order to retrieve the intrinsic color.
Human eyes are good at judging the intrinsic color under different light sources
because of the color constancy being memoried by the biological vision system (in the V4 cells~\cite{khalil2015biologically}).
HVS involves both photo-receptors and neurons in eyes,
and contains additional complex processing in our brain.
We memorize what objects are supposed to be in color under different environments,
and perform corresponding adjustment subconsciously.
However, digital cameras often have difficulty on performing similar adjustment, \textit{i.e.} ``Auto White Balance" (AWB).
This is mainly caused by the illuminant tint influence, which usually affects image sensor directly.
An accurate illuminant estimation method, aiming at correct compensation for photosensitive devices, is needed to handle wide range of light conditions.
The old-school style of camera AWB is implemented in a similar way as HVS,
in the form of ``matrix metering" by recording scene information into a prepared database.
% More specifically, this kind of approaches try to label the calibrations on image sensors according to
% a limited number of light sources in advance before delivery (\textit{i.e.} color temperature calibration).
However, the traditional way involves a lot of manual efforts and performs badly in complex illumination environment.
For example, indoor environment usually has multi-illuminant with complicated reflections, which
makes it almost impossible to be labeled.

\section{Related Work}
% Old school mathematic method on color constancy problem
There are many traditional methods try to determine the color of objects under the canonical illuminant
with specific assumptions.
Different assumptions have given rise to numerous illuminant estimation methods that
can be divided into two main groups.
First of these groups contains low-level statistic based methods, such as
White-patch~\cite{brainard1986analysis},
Gray-world~\cite{buchsbaum1980spatial},
Shades-of-Gray~\cite{finlayson2004shades}, Grey-Edge (1st and 2nd order)~\cite{van2007edge},
using the bright pixels~\cite{joze2012role}, \textit{etc}.
The second group includes machine-learning based methods, such as gamut mapping (pixel, edge, and intersection based)~\cite{barnard2000improvements},
natural image statistics~\cite{gijsenij2011color}, Bayesian learning~\cite{gehler2008bayesian},
spatio-spectral learning (maximum likehood estimation)~\cite{chakrabarti2012color}, using color/edge moments~\cite{finlayson2013corrected},
using regression trees with simple features from color distribution~\cite{cheng2015effective} and
performing various kinds of spatial localizations~\cite{barron2015convolutional},~\cite{DBLP:journals/corr/BarronT16}.

These assumptions based methods~\cite{brainard1986analysis},
~\cite{buchsbaum1980spatial},~\cite{finlayson2004shades},~\cite{van2007edge},
~\cite{joze2012role}, \textit{etc}, usually perform badly when there are no white patches in the scene.
\textit{e.g.} scene with large area of green grass or blue sea.
The current industrial way of implementing AWB algorithms is based on combining statistic-based methods with feature learning.
The approach mainly consists of: (1) scene recognition,
(2) estimation under different scene~\cite{battiato2012instant}.

% Two major challenge (1): no white-patch (2)multi source
% Cuurent work of use CNN approach, can solve (1), limition on (2)
Because of the outstanding learning performance on various computer vision tasks based on CNN
compared with traditional machine learning methods,
the state-of-the-art single illuminant estimation techniques~\cite{bianco2015color},~\cite{lou2015color},
\cite{shi2016deep},~\cite{hu2017fc},~\cite{DBLP:journals/corr/BarronT16},~\cite{bianco2015single},~\cite{gijsenij2012color}
harness the power of CNN to train illuminant estimation models using large training sets
composed of minimally processed images (linear RGB image generated from RAW) and
their associated labels of single illumination vector.
\cite{bianco2015color} operates on sampled image patches as input,
uses a CNN structure consists of convolutional layer, max pooling,
fully connected layer and three output nodes for illumination vector.
\cite{lou2015color} proposes a framework using CNN to obtain more accurate light source estimation.
They reformulate illuminant estimation as a CNN-based regression problem.
\cite{shi2016deep} proposes a CNN architecture with two interacting sub-networks,
\textit{i.e.} a hypotheses network (HypNet) and a selection network (SelNet) for better performance.
\cite{hu2017fc} presents a fully convolutional network architecture with confidence-pooling layer which
is able to determine ``what to learn" and ``how to pool" automatically from training data.
FFCC~\cite{DBLP:journals/corr/BarronT16} reformulates the problem of color constancy as
a 2D spatial localization task in a log-chrominance space,
and proceeds Bivariate Von Mises (BVM) estimation procedure on the frequency domain.
Especially, they use camera meta-data (\textit{e.g.} shutter speed) as extra semantic features to
train a CNN model with better performance.

Other techniques like~\cite{bianco2015color},~\cite{lou2015color},~\cite{shi2016deep},~\cite{hu2017fc}, \textit{etc} aim at
using deep CNN to extract high level feature for illuminant estimation,
and they consider CNN as a ``black box" global mapping function directly map the input image to its corresponding illumination vector.
In this paper, we present a CNN architecture for generating a local-based self-adaptive meaningful kernel for its own input.
The illumination vector will be calculated based on the refrence image computed by applying the learned kernel on its input image.
% One advantage of our design is that, without aiming at extracting high-level feature,
% the input size along with the CNN architecture can be more efficient.

% multi-source related work
Certainly, the ``no white patch" case can be solved by CNN based techniques because of its data-driven property.
\textit{e.g.}, if training data contain enough green grass scene,
the model will learn to predict correctly for similar green scene.
However, more often than not, the assumption of a uniform illuminant is an insufficient approximation under
real-world illumination conditions (multiple light sources, shadows, interreflections, \textit{etc}).
\cite{riess2011illuminant} presents a physics-based approach for illuminant color estimation of arbitrary images,
which is explicitly designed for handling images with multiple illuminants.
Its building block is the computation of statistically robust local illuminant estimates which
are then used in deriving the number and color of the dominant illuminants.
\cite{beigpour2014multi} formulates the multi-illuminant estimation problem as an energy minimization task within
a Conditional Random Field over a set of local illuminant estimates.
\cite{bianco2015single} and~\cite{gijsenij2012color} propose a similar methodology to extend existing algorithms
by producing corresponding local estimation with image patches rather than using the entire image.
\cite{gijsenij2012color} combines local patch estimations based on a modified diagonal model,
\cite{bianco2015single} trains a non-linear mapping based on radial basis function kernel over
local statistics of local estimations as a local-to-global regressor.
% The parameters of the mapping are obtained by minimizes the median angular error on the training set.
Techniques similar to~\cite{riess2011illuminant},~\cite{beigpour2014multi},~\cite{gijsenij2012color},~\cite{bianco2015single}, \textit{etc}
manage multi-illuminant problem into two stages: (1) split input image into local patches,
then estimate on these patches isolatedly; (2) combine the local estiamtions in a super-pixel level.
% However, these techniques compute local estiamtions from restricted local patches,
% which we believe local estiamtions
However, local estimations make more sense when it is driven from global information,
and less accurate when only from its restricted local information.

The rest of this paper is organized as following: Section~\ref{section_framework} formalizes the illuminant estimation problem
and illustrates our proposed framework in detail.
Section~\ref{section_experiments} describes the dataset and experimental results along with evaluation.
We demonstrate the proposed method on four benchmarks and reach the largest relative improvement of
16\%
compare to the state-of-the-art method.
In Section~\ref{section_c_and_d}, we present a summary and discussion of possible future works.

\section{Proposed Framework}
\label{section_framework}
\begin{figure*}
\begin{center}
% \fbox{\rule{0pt}{3.5in} \rule{.9\linewidth}{0pt}}
\includegraphics[width=\linewidth]{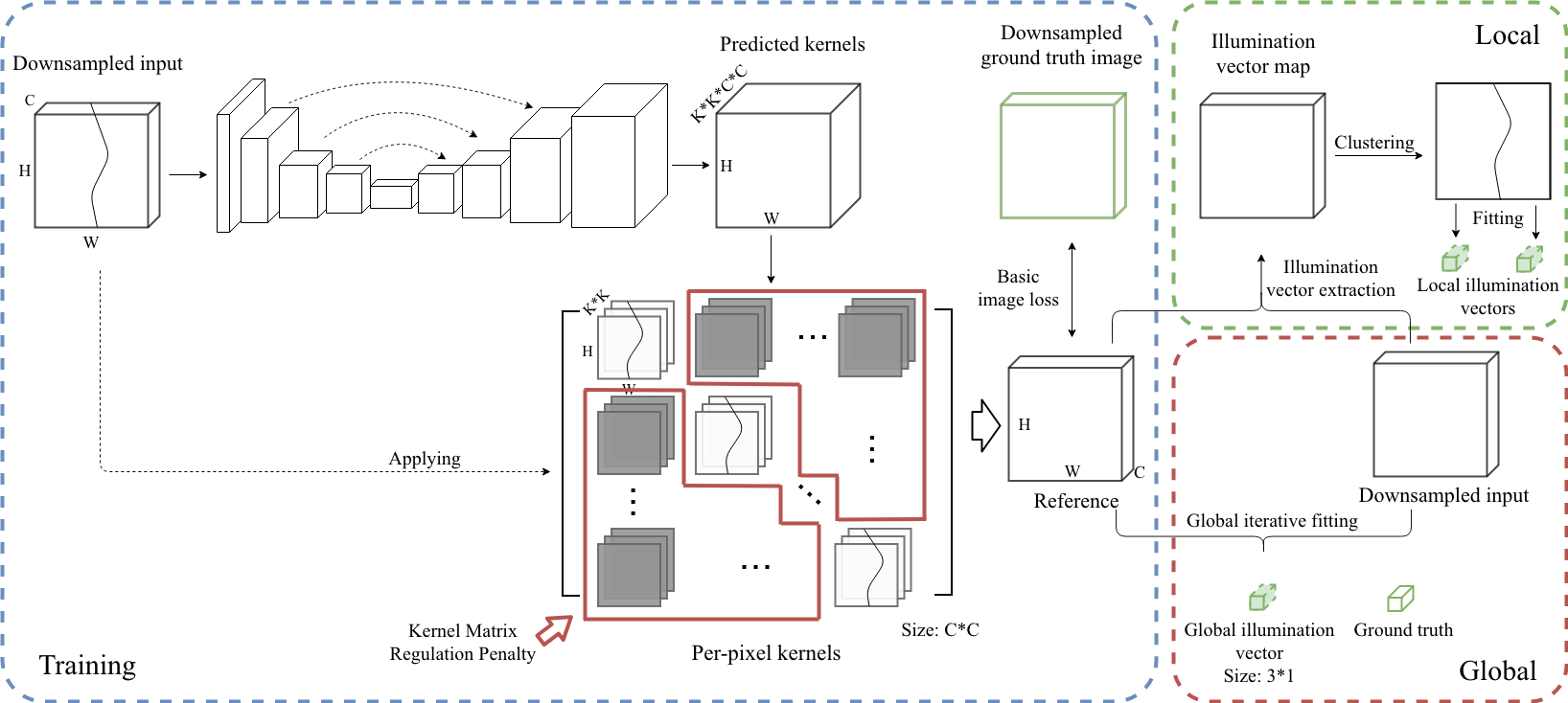}
\end{center}
   \caption{The proposed framework,
   self-adaptive kernels prediction by CNN encoder-decoder architecture with skip connections.
   Applying extracted kernels on downsampled input image results to a reference,
   which further used in iterative fitting globally and locally to compute illumination vector(s).
   % Final illuminant estimation in the original resolution is computed by applying the illumination vectors.
   }
\label{fig:framework}
\end{figure*}

% problem define , matrix multiple
In this paper, we present a framework to estimate single and multi-illuminant based on fully convolutional neural network,
local illuminant is estimated from the whole image instead of patches as in works like
~\cite{riess2011illuminant},~\cite{beigpour2014multi},~\cite{gijsenij2012color},~\cite{bianco2015single}, \textit{etc}.
Also different from other works~\cite{bianco2015color},~\cite{lou2015color},~\cite{shi2016deep},~\cite{hu2017fc}, \textit{etc}
which directly map input image into its corresponding illumination vector, we use CNN to learn an explicable kernel.
The proposed global-to-local aggregation framework (see Figure~\ref{fig:framework}) includes:
\begin{enumerate}
  \setcounter{enumi}{0}
  \item
  % \item Instead of using CNN as a ``black box'' mapping function,
  An encoder-decoder CNN structure to learn local self-adaptive kernels from global information.
  \item Confidence measurement based on the learned kernels and its input image.
  \item Clustering based iterative fitting for computing single and multi-illumination vectors subregionally.
\end{enumerate}
Each of these three stages will be explained in detail in the following subsections.

% ******big figue*******

\subsection{Problem Formulation}
The aim of illuminant estimation is to transform image captured under unknown light sources
into a color-balanced one, as if they appear under the canonical illuminant.
Usually, this is done by using a diagonal model:
\begin{equation}
I^c = \Lambda^{u,c}I^{u}
\label{eq:illu}
\end{equation}
where $I^{u}$ is the image taken under unknown illuminant environment,
$I^{c}$ is the transformed image appears as if it is taken under the canonical illuminant.
The $\Lambda^{u,c}$ is the transform diagonal matrix which takes the following form:

\begin{equation}
\Lambda^{u,c} =
\left[
\begin{array}{ccc}
\alpha & 0 & 0\\
0 & \beta & 0\\
0 & 0 &\gamma
\end{array}
\right]
=
\left[
\begin{array}{ccc}
\frac{L_R^c}{L_R^{u}} & 0 & 0\\
0& \frac{L_G^c}{L_G^{u}} & 0\\
0 & 0 & \frac{L_B^c}{L_B^{u}}
\end{array}
\right]
\end{equation}
where $L^u$ is unknown illuminant and $L^{c}$ is canonical illuminant.
For AWB module in camera pipeline, $\frac{L_R^c}{L_R^{u}}$, $\frac{L_G^c}{L_G^{u}}$, $\frac{L_B^c}{L_B^{u}}$,
usually alias $R_{gain}$, $G_{gain}$, $B_{gain}$ respectively.

AWB requires an automatically channel wised correction based on the prediction of illuminant color.
Since here we might need different gain values for different objects or different illuminant situations,
we expect CNN structure to tackle this problem.

\subsection{Self-adaptive Kernel Learning}
The design of proposed CNN network architecture aims to directly use the pixel-wise information from the input image,
with the minor concern of extracting the high-level feature.
In this paper, we use the CNN architecture in encoder-decoder design with skip connections.
Such architecture has been proved to be efficient in various pixel-level applications.
% \textit{e.g.}
For example,
``U-net" structure for image segmentation~\cite{DBLP:journals/corr/RonnebergerFB15},
image to image pixel wise translation~\cite{isola2017image},
image denoising from a burst of input images~\cite{DBLP:journals/corr/abs-1712-02327}.

Unlike works~\cite{DBLP:journals/corr/RonnebergerFB15},~\cite{isola2017image} try to
synthesize the output image pixels using this architecture,
in this paper, we use it for learning local self-adaptive kernels from global information.
Instead of directly using CNN to predict illumination vector,
we tackle this problem by first learn a intermediate mutable function \textit{i.e.} self-adaptive kernels,
final prediction of illumination vector will be computed based on learned spatial variant kernels and corresponding input.
Such global-to-local aggregation brings significant improvement to the single illuminant estimation
and better addresses multi-source illumination problem as well as extracting confidence measurement.

% The predicted kernels usually outperform traditional kernels because it computes spatially different kernels
% rather than one kernel for all image locations.
% For example, in the case of burst image denoising~\cite{DBLP:journals/corr/abs-1712-02327},
% for an input image with $C$ channels (\textit{e.g.} $C$ images in a burst) to output a single channel image,
% a kernel map with $K^2C$ channels is extracted, which then reshaped into $C$ linear filters with size of $K^2$,
% and applied into corresponding pixel locations to get the reference output.

Considering the usage of channel-wise correction in WB task,
we first try predicting kernel maps (with the same size as downsampled input image) separately for each channel:
$C$ kernel maps ($C=3$ for a single RGB image) with $K^2$ (K-order, $K{\times}K$ kernel) channels for each pair of input-output channels.
The value at each pixel $p$ in each output channel $Y_c$ of reference image will be computed as pixel-wise dot product as following:
\begin{equation}
Y^p_c = <f^p_c, V^p(X_c)>
\end{equation}
where $V^p(X)$ denotes the $K\times{K}$ neighbourhood of pixel $p$ in each input channel $X_c$, and $f^p_c$ is its corresponding kernel.
Such design of spatial adaptive kernel aims at enhancing local illuminant estimation from global information
instead of treating image patches independently (Figure~\ref{fig:framework}).

Moreover, in order to make fully use of image information,
% However, such channel-separate kernel prediction did not give outstanding performance in our experiments.
% which indicates that we may need cross-channel information and operations.
we propose computing additional kernels for cross-channel operations,
which yields better performance in our experiments.
We compute kernel maps with $K^2\times{C^2}$ channels, and the reference output
$Y$ is obtained by applying the learned kernels across all channels:
\begin{equation}
Y^p_c = \sum_{i=1}^{C}<f^p_{c,i}, V^p(X_i)>
\label{eq:kpn}
\end{equation}
This adjustment favors the learning in two ways:
(1) It increases the model complexity without deepening the network structure.
(2) It compensates the insufficient optimization when channel wise-information is not rich enough.
This cross-channel approach achieve the best results in all following experiments.

The loss function we used to learn the kernels consists of two parts:
the basic image loss similar to the loss function in~\cite{DBLP:journals/corr/abs-1712-02327},
and the additional kernel matrix regulation penalty to encourage kernels to utilize corresponding channel information more.

The basic image loss is a weighted average of L2 distance on pixel intensities and
L1 distance on pixel gradients as compared to the ground truth image,
which is obtained by applying accordingly ground truth illumination vector on the input image.
We apply the loss after applying the sRGB transfer function for gamma correction,
which can produce a more perceptually relevant estimation~\cite{DBLP:journals/corr/abs-1712-02327}.
The basic L1 and L2 image loss are shown as:
\begin{equation}
\left.\begin{aligned}
\ell_1(\hat Y, Y^*) & = ||\nabla \Gamma(\hat Y) - \nabla \Gamma(Y^*)||_1 \\
\ell_2(\hat Y, Y^*) & = ||\Gamma(\hat Y) - \Gamma(Y^*)||_2
\end{aligned}\right.
\label{img_loss}
\end{equation}

Where $\nabla$ is the finite difference operator that convolves its input with $[-1,1]$ and $[-1,1]^T$,
and $\Gamma$ denote for the sRGB transformation for both input and output channel:
\begin{equation}
\Gamma(X)={\begin{cases}12.92X,&X\leq 0.0031308\\(1+a)X^{1/2.4}-a,&X>0.0031308\end{cases}}
\label{rgb2srgb}
\end{equation}
where $a = 0.055$ and $X$ denotes for linear input channel $R$, $G$, or $B$.

Apart from image loss, we introduce L1 kernel matrix regulation penalty during training (Equation~\ref{filt_reg}).
This penalty punishes the kernels that do not align with the same input and output channel.
The intuition is: although we utilize cross-channel information,
we expect the learned kernels $F_{i,j}$ still focus on the corresponding channel
due to the channel-wise property persevering in WB training dataset.
The proposed L1 kernel matrix regulation penalty forms as following:
\begin{equation}
\left.\begin{aligned}
R(F) & = \sum_{i,j}w_{i,j}||F_{i,j}||_1 \\
& where \ w_{i,j} = {\begin{cases}0 &  i = j \\
   1 & i != j \end{cases}}
\end{aligned}\right.
\label{filt_reg}
\end{equation}

Our final loss function forms as following:
\begin{equation}
\left.\begin{aligned}
\textbf{L}(X,Y^*) = \lambda_1 \ell_1 & (f(X),Y^*) + \lambda_2 \ell_2(f(X),Y^*) + \lambda_3 R(F) \\
& where \ f(X) = \sum_{j}X_i \cdot F_{i,j}
\end{aligned}\right.
\label{whole_loss}
\end{equation}
where $f$ denotes for applying the learned kernel on the input image as Equation~\ref{eq:kpn},
loss weights of $\lambda_1$, $\lambda_2$ and $\lambda_3$ are set to $1.0, 1.0, 0.01$ respectively in our experiments.

By applying learned kernels to downsampled input image, we get a reference image.
The final illumination vector is produced by iterative fitting using the reference image as described in subsection~\ref{subsection_iterative_fitting}.
In addition, prediction confidence can be derived by learned kernels (subsection~\ref{subsection_confidence_map}),
thereby the proposed method offers much more freedom for further hand-engineered calibration
as well as following stages in camera image processing pipeline.

\subsection{Confidence Estimation}
\label{subsection_confidence_map}
As mentioned before,
we predict $C\times{C}$ kernels from downsampling resized input image.
Such ``kernel matrix" form enables that each $C$ input channels can contribute to output channels.
Due to the channel-wise property in WB task as well as our kernel matrix regulation penalty,
we define the confidence value to be the rate between the ``contribution'' from the corresponding input channel and the others.

Confidence value for each channel $c$ in pixel $p$ is defined as following:
\begin{equation}
Conf_c^p \propto \frac{<|f^p_{c,c}|, V^p(X_c)>}{\sum_{i=1}^{C}<|f^p_{c,i}|, V^p(X_i)>}
\label{eq:conf_def}
\end{equation}

By combining confidence across all channels, we can get a uniformed estimation.
The empirical evidence shows that confidence value is proportional to the mean of all confidence maps,
and inversely proportional to the variance of all confidence maps:\begin{equation}
Conf \propto \frac{1}{C}\sum_{c=1}^{C} \frac{mean(Conf_c) + \epsilon}{var(Conf_c) + \epsilon}
\label{eq:compute_conf}
\end{equation}

\begin{figure*}[tb]
\begin{center}
% \fbox{\rule{0pt}{3in} \rule{.9\linewidth}{0pt}}
\includegraphics[width=\linewidth]{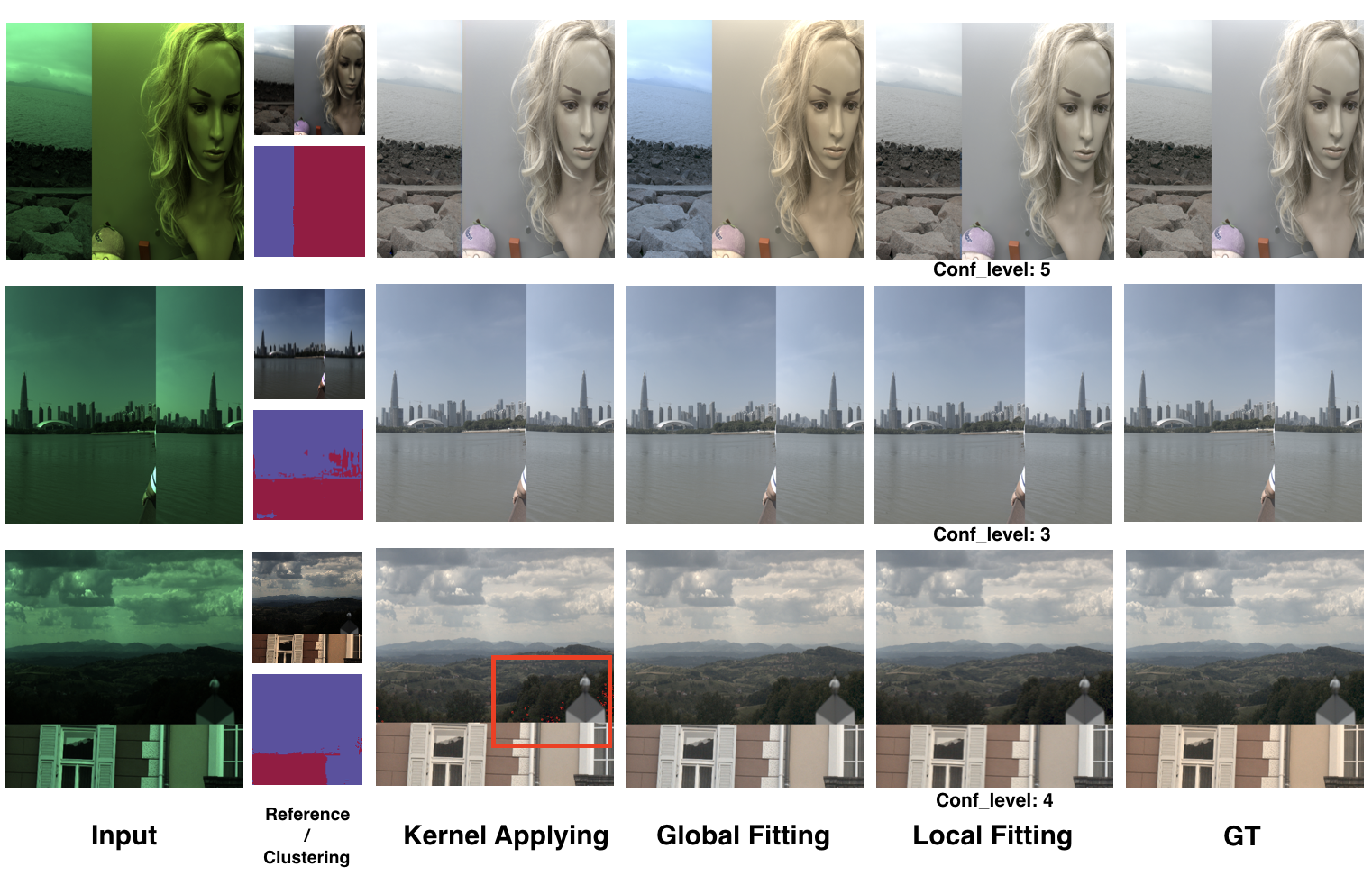}
\end{center}
\caption{Examples of three typical cases.
        \textbf{Input:} Input image in original resolution.
        \textbf{Reference:} The reference image computed by applying kernels on downsampled input.
        \textbf{Clustering:} The illuminant cluster mask on computed illumination vector map.
        \textbf{Kernel Applying:} Image in original resolution computed using illumination vector map.
        \textbf{Global Fitting:} Output image using global iterative fitting.
        \textbf{Local Fitting:} Output image using iterative fitting on each illuminant cluster.
        \textbf{GT:} Ground truth image generated by ground truth illumination vector.
        % \textbf{First row:} Correct clustering at boundaries.
        % \textbf{Second row:} Seperate one ligth source into two different regions.
        % \textbf{Third row:} Fail to perfectly part two regions under different light sources,
        %   though, the results is acceptable.
            }

\label{fig:visualization}
\end{figure*}

\subsection{Iterative Fitting and Illuminant Clustering}
\label{subsection_iterative_fitting}
We adopt clustering and iterative fitting to solve illumination vector subregionally.
The goal of iterative fitting is to find optimal gain values to make the white balanced image close to the reference in general.
The segmental workflow benifits in two parts compare with directly applying kernels on original input image:
First, for high-resolution images (\textit{e.g.} 4K images), the downsample resizing operation save the hardware cost;
Second, the iterative fitting method will compress the artifacts occur in underexposure region
due to the spikes in illuminant vector map.
% The spikes are caused by divided by very small value.
The example of artifacts coming from directly applying kernels is shown in Figure~\ref{fig:visualization}.
In the column of kernel applying,
the artifacts appear at the over dark region of lower right corner (last row).

% In our experiments, we minimize the L2 distance between the image $Y_g$ processed by the gain value to be optimized
% as Equation~\ref{eq:illu} given reference image $Y_{ref}$:
In WB task, the fitting target is three dimensional vector (Equation~\ref{eq:illu}).
We minimize the L2 distance between the reference image $Y_{ref}$ and the fitting results image $Y_g$:
\begin{equation}
D = ||Y_g-Y_{ref}||_2
\label{eq:fitting}
\end{equation}
where $Y_g$ is computed using the fitting target.

Due to the limited complexity of fitting target
we used downhill simplex algorithm~\cite{nelder1965simplex} with the initialization gain value as constant 1.0 in our experiment,
which works well in our evaluations.
Further improvement may be acquired by using more powerful optimization methods.

However, uniform illuminant assumption is insufficient for real-world applications.
Instead of using global fitting, we divide the input image into subregions that vary by different illuminants.
Specifically, for each input-reference pair, an illumination vector map could be acquired by the division between reference and input,
the clustering is performed on the illumination vector map.

Considering the specific of illuminant clustering (few cluster number, non-flat geometry),
we select Spectral Clustering (SC)~\cite{stella2003multiclass} in our following experiments.
SC requires the number of clusters to be specified,
in our experiments we set to 2.
% tested various cluster numbers and 2 clusters work well in our non-synthesized test set.
For two clusters, it solves a convex relaxation of the normalized cuts problem on the similarity graph:
cutting the graph in two so that the weight of the edges cut is small compared to the weights of the edges inside each cluster.
In our case, graph vertices are defined as each pixel on the illumination vector map,
and edges are defined as the similarity between two illumination vector modeled by  RBF kernel:
$
E_{ij}=exp(-\frac{||I_i - I_j||}{\sigma^2})
$
, where $\sigma$ is the parameter to control kernel variance, and the distance between $I_i$ and $I_j$ is computed as vectors angle distance:
\begin{equation}
||I_i - I_j|| = arccos(I_i \cdot I_j)
\label{eq:acos}
\end{equation}

In order to show the adaptive performance on multi-illuminance circumstance,
we concatenate two image parts that under different illumination into one image.
The clustering and fitting result on the concatenated images are shown in Figure~\ref{fig:visualization} at local fitting column.
From the result, we could see improvements on local illuminant estimation compare to global fitting.
% In the following sections, we introduce our experimental settings and results in details.

\section{Experimental Results}
\label{section_experiments}
We present the implementation and training detail in Section~\ref{subsection_impl_train}.
Quantitative evaluations of the proposed method on benchmarks are given in Section~\ref{subsection_benchmarks},
including open source benchmarks (Gehler~\cite{gehler2008bayesian},
NUS-8~\cite{cheng2014illuminant}
and one self-collected dataset captured by Sony DSC-RX100M3 single-lens reflex(DSLR) camera.
See supplement for more experiments and results.
A gamma correction of $\gamma = 1/2.2$ on linear RGB images is applied for display.
Further detail is presented in Section~\ref{subsection_detail}.

\subsection{Implementation and Training}
\label{subsection_impl_train}
Our proposed network architecture is implemented by Tensorflow~\footnote{https://www.tensorflow.org/},
network architecture is shown in Figure~\ref{fig:framework},
ReLu is used for non-linear activation except for the final layer.
We deploy variants of network architecture with different input size and filter size
for evaluating our method, our best results is from structure of
with input size $128\times128\times3$, filter size $5\times5\times3\times3$.
Moreover, benefit from our efficient design of network architecture,
the architecture with input size $64\times64\times3$, filter size $1\times1\times3\times3$ and $3\times3\times3\times3$ also
exhibit the comparable performance which require much less computational cost (see supplement),
\textit{e.g.}~\cite{hu2017fc} uses $512\times512\times3$ as input.
We apply the following data augmentation:
Randomly crop of the images, which are initially obtained by
first randomly choosing a side length that is 0.1 to 0.9 times the shorter edge of the original image;
Randomly rotation between -60 to 60 degree;
Randomly left-right top-down flip with a probability of 0.5;
Randomly divide and concatenate two input image together for one new training sample.
After augmentation, image will be resized into according input size.
We use Adam~\cite{kingma2014adam} optimization algorithm with batch size of 32 and a base learning rate of $1e^{-4}$.
The training runs one million iterations on an NVIDIA 1080Ti GPU and takes 3 days in our setting.
Our code will be made available publicly.

\begin{table}[b]
\resizebox{\columnwidth}{!}{
\begin{tabular}{l|llllll}
Method & Mean & Med. & Tri. &
\begin{tabular}{@{}c@{}} Best \\ 25\% \end{tabular} &
\begin{tabular}{@{}c@{}} Worst \\ 25\% \end{tabular} &
G.M. \\
\hline
White-Patch~\cite{brainard1986analysis}&  7.55 & 5.68 & 6.35 & 1.45 & 16.12 & 5.76\\
Gray-World~\cite{buchsbaum1980spatial}&  6.36 & 6.28 & 6.28 & 2.33 & 10.58 & 5.73\\
Edge-based Gamut~\cite{barnard2000improvements}& 6.52 & 5.04 & 5.43 & 1.90 & 13.58 & 5.40\\
1st-order Gray-Edge~\cite{van2007edge}& 5.33 & 4.52 & 4.73 & 1.86 & 10.03 & 4.63\\
2nd-order Gray-Edge~\cite{van2007edge}& 5.13 & 4.44 & 4.62 & 2.11 & 9.26 & 4.60\\
Shades-of-Gray~\cite{finlayson2004shades} & 4.93 & 4.01 & 4.23 & 1.14 & 10.20 & 3.96\\
Bayesian~\cite{gehler2008bayesian}& 4.82 & 3.46 & 3.88 & 1.26 & 10.49 & 3.86\\
General Gray-World~\cite{barnard2002comparison}& 4.66 & 3.48 & 3.81 & 1.00 & 10.09 & 3.62\\
Spatio-spectral Statistic~\cite{chakrabarti2012color}& 3.59 & 2.96 & 3.10 & 0.95 & 7.61 & 2.99\\
Interesection-based Gamut~\cite{barnard2000improvements} & 4.20 & 2.39 & 2.93 & 0.51 & 10.70& 2.76\\
Pixels-baed Gamut~\cite{barnard2000improvements}& 4.20 & 2.33 & 2.91 & 0.50 & 10.72 & 2.73\\
Cheng et al.2014~\cite{cheng2014illuminant}& 3.52 & 2.14 & 2.47 & 0.50 & 8.74 & 2.41\\
Exemplar-based~\cite{joze2014exemplar}& 2.89 & 2.27 & 2.42 & 0.82 & 5.97 & 2.39\\
Corrected-Moment~\cite{finlayson2013corrected}& 2.86 & 2.04 & 2.22 & 0.70 & 6.34 & 2.25\\
Bianco CNN~\cite{bianco2015color}& 2.63 & 1.98 & 2.10 & 0.72 & 3.90 & 2.04\\
Cheng et al. 2015~\cite{cheng2015effective}& 2.42 & 1.65 & 1.75 & 0.38 & 5.87 & 1.73\\
CCC~\cite{barron2015convolutional}& 1.95 & 1.22 & 1.38 & 0.35 & 4.76 & 1.40\\
DS-Net~\cite{shi2016deep}& 1.90 & 1.12 & 1.33 & 0.31 & 4.84 & 1.34\\
SqueezeNet-FC4~\cite{hu2017fc}& 1.65 & 1.18 & 1.27 & 0.38 & 3.78 & 1.22\\
FFCC~\cite{DBLP:journals/corr/BarronT16}& 1.61 & 0.86 & 1.02 & 0.23 & 4.27 & 1.07\\
\hline
Our method & \textbf{1.35} & \textbf{0.76} & \textbf{0.98} & \textbf{0.22} & \textbf{3.76} & \textbf{0.93} \\
\hline
\end{tabular}
}
\caption{Results on the Gehler dataset.
For each evaluation metric, the best result is highlighted with bold type.}
\label{table_gehler}
\end{table}

\subsection{Benchmark Evaluation}
\label{subsection_benchmarks}
% Sum up outperform state of art, and how extra raw to linear png

We evaluate the proposed framework on three open source benchmarks: Gehler~\cite{gehler2008bayesian},
NUS-8~\cite{cheng2014illuminant} and Cube~\cite{DBLP:journals/corr/abs-1712-00436} (see supplement for Cube dataset).
The Gehler dataset contains 568 images with both indoor and
outdoor shots taken by two high quality digital DSLR cameras (Canon 5D and Canon1D) with all settings in auto mode.
482 images for camera Canon 5D and 84 images for camera Canon 1D.
NUS-8 dataset contains total number of 1736 images captured by 8 different cameras: Canon EOS-1Ds Mark III,
Canon EOS 600D, Fujifilm X-M1, Nikon D5200, Olympus E-PL6, Panasonic Lumix DMC-GX1, Samsung NX2000, and Sony SLT-A57.
There are roughly equal numbers of images for each camera.
We split the datasets into 5 folds, and perform cross evaluating on four of five for training and remaining for testing.
Several standard metrics are reported in terms of angular distance with ground truth (see Equation~\ref{eq:acos}, in degree) :
mean, median, tri-mean of all the errors,
mean of the lowest 25\% of errors, and mean of the highest 25\% of errors.

For images in Gehler and NUS-8 datasets, a Macbeth ColorChecker (MCC) is presented for obtaining the ground truth illumination vector.
The ground truth is obtained from the difference between the two  brightest achromatic patches
containing no saturated value to get rid of the effects from the dark level and saturation level of the sensor.
In order to use those images, dark level has to be subtracted from the original images which
is provided in the ground truth files from these datasets.
Table~\ref{table_gehler} and Table~\ref{table_nus8} summarize the results of
our proposed method based on global fitting and previous algorithms on Gehler and NUS-8 dataset.
Notice that we reach the relative improvement around 16\% on Gehler dataset compare to state-of-the-art algorithm~\cite{DBLP:journals/corr/BarronT16},
and small improvement on NUS-8 datasets.
We argue the reason is that the diverse optical characteristic among different camera
results in making the distribution of input data dispersive which increase learning difficulty inevitably.
The supplement contains additional results on Cube~\cite{DBLP:journals/corr/abs-1712-00436} dataset.

\begin{table}[tb]
\resizebox{\columnwidth}{!}{
\begin{tabular}{l|llllll}
Method & Mean & Med. & Tri. &
\begin{tabular}{@{}c@{}} Best \\ 25\% \end{tabular} &
\begin{tabular}{@{}c@{}} Worst \\ 25\% \end{tabular} &
G.M. \\
\hline
White-Patch~\cite{brainard1986analysis}& 10.62 & 10.58 & 10.49 & 1.86 & 19.45 & 8.43 \\
Edge-based Gamut~\cite{barnard2000improvements}& 8.43 & 7.05 & 7.37 & 2.41 & 16.08 & 7.01 \\
Pixels-baed Gamut~\cite{barnard2000improvements}& 7.70 & 6.71 & 6.90 & 2.51 & 14.05 & 6.60 \\
Interesection-based Gamut~\cite{barnard2000improvements}& 7.20 & 5.96 & 6.28 & 2.20 & 13.61 & 6.05 \\
Gray-World~\cite{buchsbaum1980spatial}& 4.14 & 3.20 & 3.39 & 0.90 & 9.00 & 3.25 \\
Bayesian~\cite{gehler2008bayesian}& 3.67 & 2.73 & 2.91 & 0.82 & 8.21 & 2.88 \\
Natural Image Statics~\cite{gijsenij2011color}& 3.71 & 2.60 & 2.84 & 0.79 & 8.47 & 2.83\\
Shades-of-Gray~\cite{finlayson2004shades} & 3.40 & 2.57 & 2.73 & 0.77 & 7.41 & 2.67\\
General Gray-World~\cite{barnard2002comparison}&  3.21 & 2.38 & 2.53 & 0.71 & 7.10 & 2.49\\
2nd-order Gray-Edge~\cite{van2007edge}& 3.20 & 2.26 & 2.44 & 0.75 & 7.27 & 2.49\\
Bright Pixels~\cite{joze2012role}& 3.17 & 2.41 & 2.55 & 0.69 & 7.02 & 2.48\\
1st-order Gray-Edge~\cite{van2007edge}& 3.20 & 2.22 & 2.43 & 0.72 & 7.36 & 2.46\\
Spatio-spectral Statics~\cite{chakrabarti2012color} & 2.96 & 2.33 & 2.47 & 0.80 & 6.18 & 2.43\\
Corrected-Moment~\cite{finlayson2013corrected}& 3.05 & 1.90 & 2.13 & 0.65 & 7.41 & 2.26\\
Color Tiger~\cite{DBLP:journals/corr/abs-1712-00436}& 2.96 & 1.70 & 1.97 & 0.53 & 7.50 & 2.09 \\
Cheng et al. 2014~\cite{cheng2014illuminant}& 2.92 & 2.04 & 2.24 & 0.62 & 6.61 & 2.23\\
Color Dog~\cite{banic2015color}& 2.83 & 1.77 & 2.03 & 0.48 & 7.04 & 2.03 \\
DS-Net~\cite{shi2016deep}& 2.24 & 1.46 & 1.68 & 0.48 & 6.08 & 1.74\\
CCC~\cite{barron2015convolutional}& 2.38 & 1.48 & 1.69 & 0.45 & 5.85 & 1.74 \\
SqueezeNet-FC4~\cite{hu2017fc}& 2.23 & 1.57 & 1.72 & 0.47 & 5.15 & 1.71 \\
Cheng et al. 2015~\cite{cheng2015effective}& 2.18 & 1.48 & 1.64 & 0.46 & 5.03 & 1.65 \\
FFCC~\cite{DBLP:journals/corr/BarronT16}& 1.99 & \textbf{1.31} & \textbf{1.43} & \textbf{0.35} & 4.75 & 1.44\\
\hline
Our method & \textbf{1.96} & 1.41 & 1.44 & \textbf{0.35} & \textbf{4.29} & \textbf{1.36} \\
\hline
\end{tabular}
}
\caption{Results on the NUS-8 dataset.
For each evaluation metric, the best result is highlighted with bold type.}
\label{table_nus8}
\end{table}

In addition to the open source benchmarks, we provide evaluation on a self-collected dataset named MIO (massive-indoor-outdoor) dataset.
It contains massive indoor and outdoor images (over 13k) captured by Sony DSC-RX100M3 camera.
The ground truth is obtained based on the meta-file generated by DSC-RX100M3 camera.
The goal of collecting such a dataset aims at evaluating the performance of an AWB algorithm to
the maximum close to real world application instead of making benchmark for training with accurate ground truth.
We split the training, validation and test sets as 9k, 2k, 2k images.
We only do minimal processing on the resolution of 4864x3648 (4:3) raw images to get linear training images as same as on
Gehler~\cite{gehler2008bayesian} and NUS-8~\cite{cheng2014illuminant} datasets.
The dataset contains both outdoor images and indoor images in parts of Tokyo and ShengZhen under various seasons.
Especially, it contains multiple corner case scene which traditional camera algorithm performs badly on.
\textit{e.g.} full scene with green grass, human skin in indoor (2500-3000K).
Table~\ref{table_sony} exhibits the performance of proposed methods and other methods on MIO dataset,
we outperform state-of-the-art algorithm in all the evaluation metrics.

\begin{table}[b]
\resizebox{\columnwidth}{!}{
\begin{tabular}{l|llllll}
Method & Mean & Med. & Tri. &
\begin{tabular}{@{}c@{}} Best \\ 25\% \end{tabular} &
\begin{tabular}{@{}c@{}} Worst \\ 25\% \end{tabular} &
G.M. \\
\hline
White-Patch~\cite{brainard1986analysis} & 5.59 & 4.42 & 5.21 &1.09 & 15.25 & 4.28\\
Gray-world~\cite{buchsbaum1980spatial}& 3.75 & 2.95 & 3.18 & 0.89 & 9.02 & 2.89 \\
Shades-of-Gray~\cite{finlayson2004shades} & 2.61 & 1.72 & 1.96 & 0.47 & 8.01 & 1.91 \\
1st-order Gray-Edge~\cite{van2007edge}&  2.47 & 1.61 & 1.89 & 0.43 & 6.32 & 1.90 \\
2nd-order Gray-Edge~\cite{van2007edge}& 2.46 & 1.59 & 1.91 & 0.45 & 6.19 & 1.89 \\
Bright Pixels~\cite{joze2012role} & 2.42 & 1.57 & 1.84 & 0.46 & 5.99 & 1.89 \\
FFCC~\cite{DBLP:journals/corr/BarronT16}& 1.09 & 0.81 & 0.82 & 0.30 & 1.93 & 0.79 \\
SqueezeNet-FC4~\cite{hu2017fc}& 0.96 & 0.70 & 0.72 & 0.26 & 1.78 & 0.68\\
\hline
Our method & \textbf{0.72} & \textbf{0.59} & \textbf{0.60} & \textbf{0.23} & \textbf{1.61} & \textbf{0.53} \\
\hline
\end{tabular}
}
\caption{Results on the MIO dataset.
For each evaluation metric, the best result is highlighted with bold type.}
\label{table_sony}
\end{table}

\subsection{Detail and Examples}
\label{subsection_detail}
% and explain how to compute confidence
% [formulation]
% explain and visulization examples

In order to imitate the appearance of two distinct light sources in one image,
we randomly divide images into rectangles and concatenate them together.
We argue that synthetic test samples could appeal the ability on multi-illuminant estimation task more evidently.
Because in real environment it is inevitable that two light sources will overlap with each other which
usually do not have explicit boundaries.

Figure~\ref{fig:visualization} shows three typical results of the proposed method under two light sources:
correct clustering at boundaries;
separating one light source into two different regions;
false dividing two regions under different light sources perfectly.
\textbf{Reference} is computed based on Equation~\ref{eq:kpn}.
\textbf{Kernel Applying:} Extract a local RGB illumination vector map by using the reference image divide by input image (downsampled size),
then apply the low-resolution illumination vector map on the original input image (high resolution) based on the scale.
\textbf{Global Fitting:} Iterative fitting of Eqution~\ref{eq:fitting} under the assumption of single illuminant.
\textbf{Local Fitting:} Apply Spectral Clustering (in Section~\ref{subsection_iterative_fitting}) on local RGB illumination vector map,
then iterative fitting subregionally.
By using clustering fitting we can remove the artifacts (third row red points in the dark region) and save hardware overhead as well.

\begin{figure}[t]
\begin{center}
   \includegraphics[width=\linewidth]{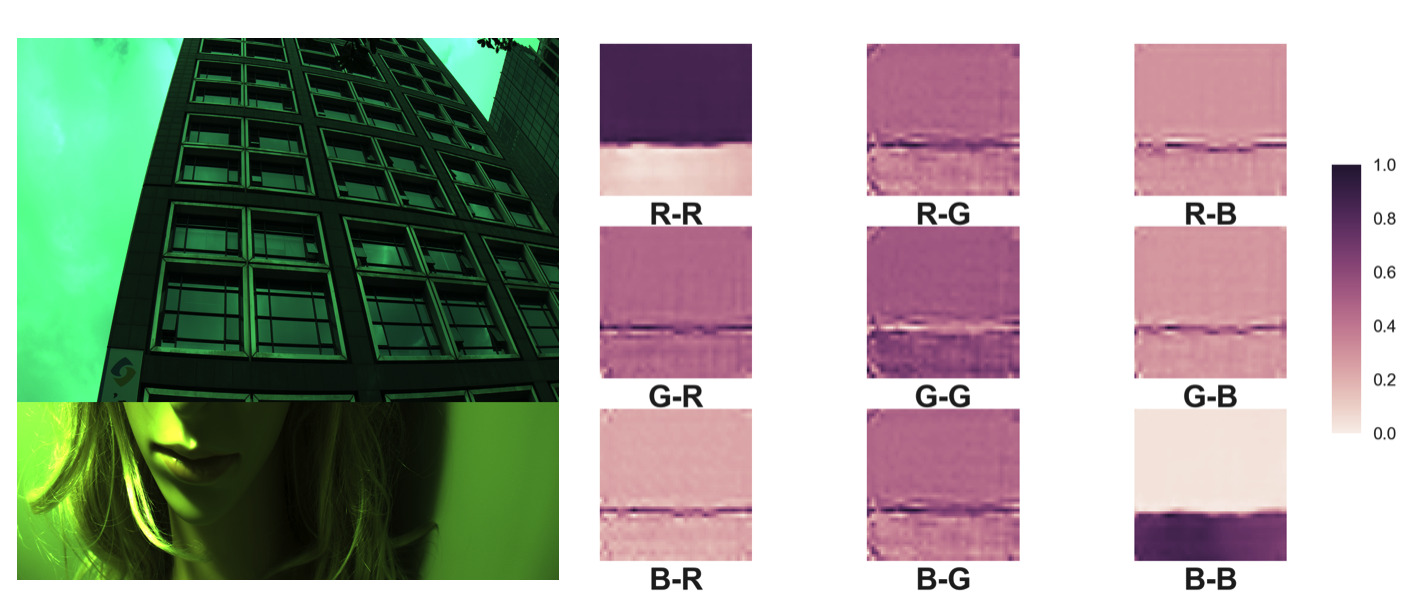}
\end{center}
   \caption{
    Visulization of learned kernels. \textbf{left}: RGB input image \textbf{right}: $3{\times}3$ kernel maps (RGB),
    value is normalized into [0,1].
    % Ground truth for top part of input image is [2.65234375, 1.0, 1.710937499999999], below is [1.4882812499999998, 1.0, 3.4218749999999996].
    Ground truth (RGB order) of top part of input image is [0.80, 0.30, 0.52], below is [0.38, 0.26, 0.89].
    Notice the kernels in diagonal (R-R, G-G, B-B) have relatively large value,
    and boundaries have less confidence.
    }
\label{fig:vis_kernel}
\end{figure}

\begin{paragraph} {Kernel visualization}
  We visualize the learned kernel in Figure~\ref{fig:vis_kernel}.
  The input image is divide and concatenated by two images under distinct illuminants:
  Here we demonstrate the learned kernel from architecture 64x64x3 / 1x1x3x3 (see supplement).
  We denote the filter's $a$ channel used for computing the reference's $b$ channel as $a$-$b$.
  The success of learning shows the aligned kernels (R-R, G-G, B-B) have larger values than the others in Figure~\ref{fig:vis_kernel}.
  From the visualization, we can see that prediction will have less confidence on the boundaries (Equation~\ref{eq:conf_def}).
\end{paragraph}

\begin{paragraph} {Confidence extraction}
  Considering AWB in camera image processing pipeline, $R_{gain}$ and $B_{gain}$ are normalized with $G_{gain}$,
  instead of computing the uniform confidence based on all channels (Equation~\ref{eq:compute_conf}),
  we practically calculate the confidence on R and B channel as:
  \begin{equation}
    Conf = \frac{\mu_R +\mu_B}{|\mu_R - \mu_B| \cdot \sqrt{{\mu_R}^2 + {\mu_B}^2}}
    \label{eq:confi}
  \end{equation}
  where $\mu_R$ and $\mu_B$ denote for the mean of $Conf_R$ and $Conf_B$ respectively.
  Equation~\ref{eq:confi} indicates we expect the mean of confidence of R and B channel to be large and balanced as well.
  In order to make the confidence estimation more robust and easier to be combined with other algorithms,
  we separate confidence into five levels (1-5), which span evenly on the range of confidence on training set.
  Higher level means higher confidence.
  \textit{i.e.} Level 5 is the confidence value is bigger than 0.8 times mean confidence value of all training samples.
  See supplement for more results.
\end{paragraph}

% deprecated
% \subsection{Stability Performance}
% \label{subsection_stability_check}
% explain dirty types: rotate, crop, clip, ISO mimic

\section{Conclusion and Discussion}
\label{section_c_and_d}
% cons
In this paper, we present self-adaptive kernel learning for single and multi-illuminant estimation.
The proposed global-to-local aggregation framework is able to predict region-based illuminant vector
(single or multiple according to the clustering results) using
the global information instead of training with image patches.
We evaluate the proposed method on open source benchmarks, additionally,
a self-made benchmark dataset with 13k images has been created for evaluation.
Experiments demonstrate our proposed method outperforms the state-of-the-art methods with the largest relative improvement of 16\%.

% extension
Typical image processing pipeline involves a lot of components which need to be manually finetuned.
The proposed framework can be easily extended to other modules in pipeline processing.
\textit{e.g.} color matrix transformation module (CMT).
Different from AWB which aims to render the acquired image as closely as possible under the canonical illuminant,
CMT transforms the image data into a standard color space.
This transformation is needed because the spectral sensitivity functions of the sensor color channels
rarely match the desired output color space.
CMT is usually performed by using a linear transformation matrix,
which could also use the same framework to learn with in a ``white box" manner.
% One challenge for adaptive CMT training is the pair-wise training sample is hard to make.
% Due to the recent success of unpair-wised learning of Generative Adversarial Networks~\cite{kim2017learning},~\cite{zhu2017unpaired},
% we can use the finely toned images as guidance for training.
% The form of the cost function will be automatically modeled by the discriminator.

The model does not generalize well among differnt cameras sources.
For Gehler dataset, experiments show a significant improvement,
though, for NUS-8 dataset (images are captured from eight cameras),
it shows less improvement due to the optical divergence of its training images.
The future work leaves the model-transfer ability to be improved.

{\small
\bibliographystyle{ieee}
\bibliography{paper}
}

\end{document}